\documentclass[11pt,a4paper]{article}
\usepackage[hyperref]{acl2019}
\usepackage{times}
\usepackage{latexsym}
\usepackage{amsmath}
\usepackage{graphicx}
\usepackage{float}

\usepackage{url}

\aclfinalcopy 

\title{JUMT at WMT2019 News Translation Task: A Hybrid approach to Machine Translation for Lithuanian to English}

\author{Sainik Kumar Mahata, Avishek Garain, Adityar Rayala,\\ \textbf{Dipankar Das, Sivaji Bandyopadhyay}\\
Computer Science and Engineering \\
	Jadavpur University, Kolkata, India \\
  {\tt sainik.mahata@gmail.com, avishekgarain@gmail.com, mailsofadityar@gmail.com},\\ {\tt dipankar.dipnil2005@gmail.com, sivaji\_cse\_ju@yahoo.com}}

\date{}

\begin{document}
\maketitle
\begin{abstract}
In the current work, we present a description of the system submitted to WMT 2019 News Translation Shared task. The system was created to translate news text from Lithuanian to English. To accomplish the given task, our system used a Word Embedding based Neural Machine Translation model to post edit the outputs generated by a Statistical Machine Translation model. The current paper documents the architecture of our model, descriptions of the various modules and the results produced using the same. Our system garnered a BLEU score of 17.6.

\end{abstract}

\section{Introduction}
\label{Intro}
\textbf{M}achine \textbf{T}ranslation (MT) is automated translation of one natural language to another using a computer. Translation, itself, is a very tough task for both humans as well as a computer. It requires a thorough understanding of the syntax
and semantics of both the languages under consideration. For producing good translations, a MT system needs good quality and sufficient amount of parallel corpus \cite{mahata:2016wmt2016, mahata:2017bucc2017}.

In the modern context, MT systems can be categorized into \textbf{S}tatistical \textbf{M}achine \textbf{T}ranslation (SMT) and \textbf{N}eural \textbf{M}achine \textbf{T}ranslation (NMT). SMT has had its share in making MT very popular among the masses. It includes creating statistical models, whose input parameters are derived from the analysis of bilingual text corpora, created by professional translators \cite{weaver:1955translation}. The
state-of-art for SMT is Moses Toolkit\footnote{http://www.statmt.org/moses/}, created by \citet{koehn:2007moses}, incorporates subcomponents like Language Model generation, Word Alignment and Phrase Table generation. Various works have been done in SMT \cite{lopez:2008statistical,koehn:2009statistical} and it has shown good results for many language pairs.

On the other hand NMT \cite{bahdanau:2014neural}, though relatively new, has shown considerable improvements in the translation results when compared to SMT \cite{mahataMTIL}. This includes better fluency of the output and better handling of the Out-of-Vocabulary problem. Unlike SMT, it doesn’t depend on alignment and phrasal unit translations \cite{kalchbrenner:2013recurrent}. On the contrary, it uses an Encoder-Decoder approach incorporating Recurrent Neural Cells \cite{cho:2014properties}. As a result, when given sufficient amount of training data, it gives much more accurate results when compared to SMT \cite{doherty:2010eye, vaswani:2013decoding, liu:2014recursive}.

For the given task\footnote{http://www.statmt.org/wmt19/translation-task.html}, we attempted to create a MT system that can translate sentences from Lithuanian to English. Since, using only SMT or NMT models leads to some or the other disadvantages, we tried to use both in a pipeline. This leads to an improvement of the results over the individual usage of either SMT or NMT. The main idea was to train a SMT model for translating Lithuanian language to English. Thereafter, a test set was translated using this model. Then, a word embedding based NMT model was trained to learn the mappings between the SMT output (in English) and the gold standard data (in English).

The organizers provided the required parallel corpora, consisting of 9,62,022 sentence pairs, for training the translation model. Among this, 7,62,022 pairs was used to train the SMT system and 2,00,000 pairs were used to test the SMT system and then train the NMT system. The statistics of the parallel corpus is depicted in \ref{Table1}.

\begin{table}[h]
\centering
\begin{tabular}{|l|l|}
\hline
\textbf{\# sentences in Lt corpus} & 9,62,022 \\ \hline
\textbf{\# sentences in En corpus} & 9,62,022 \\ \hline
\textbf{\# words in Lt corpus} & 1,16,65,937 \\ \hline
\textbf{\# words in En corpus} & 1,56,22,488 \\ \hline
\textbf{\# word vocab size for Lt corpus} & 4,88,593 \\ \hline
\textbf{\# word vocab size for En corpus} & 2,27,131 \\ \hline
\end{tabular}
\captionsetup{justification=centering}
\caption{Statistics of the Lithuanian-English parallel corpus provided by the organizers. "\#" depicts No. of. "Lt" and "En" depict Lithuanian and English, respectively. "vocab" means vocabulary of unique tokens. }
\label{Table1}
\end{table}

The remainder of the paper is organized as follows. Section \ref{sec:methdology} will describe the methodology of creating the SMT and the NMT model and will include the preprocessing steps, a brief summary of the encoder-decoder approach and the architecture of our system. This will be followed by the results and conclusion in Section \ref{sec:results} and \ref{sec:conclusion}, respectively.

\section{Methodology}
\label{sec:methdology}
\subsection{SMT}
\label{subsec:SMT}
For designing the model we followed some standard preprocessing steps on 7,62,022 sentence pairs, which are discussed below.
\subsubsection{Preprocessing}
The following steps were applied to preprocess and clean the data before using it for training our Statistical machine translation model. We used the NLTK toolkit\footnote{https://www.nltk.org/}
for performing the steps.
\begin{itemize}
\item \textbf{Tokenization}: Given a character sequence and a defined document unit, tokenization is the task of chopping it up into pieces, called tokens. In our case, these tokens were words,
punctuation marks, numbers. NLTK supports tokenization of Lithuanian as well as English texts.
\item \textbf{Truecasing}: This refers to the process of restoring case information to badly-cased or non-cased text \cite{lita:2003truecasing}. Truecasing helps in reducing data sparsity.
\item \textbf{Cleaning}: Long sentences (No. of tokens $>80$) were removed.
\end{itemize}
\subsubsection{Moses}
Moses is a statistical machine translation system that allows you to automatically train translation models for any language pair, when trained with a large collection of translated texts (parallel corpus). Once the model has been trained, an efficient search algorithm quickly finds the highest probability translation among the exponential number of choices.

We trained Moses using 7,62,022 sentence pairs provided by WMT2019, with Lithuanian as the source language and English as the target language. For building the Language Model we used KenLM\footnote{https://kheafield.com/code/kenlm/} \cite{Heafield-kenlm} with 7-grams from the target corpus. The English monolingual corpus from WMT2019 was used to build the language model

Training the Moses statistical MT system resulted in generation of Phrase Model and Translation Model that helps in translating between source-target language pairs. Moses scores the phrase in the phrase table with respect to a given source sentence and produces best scored phrases as output. 

\subsection{NMT}
Neural machine translation (NMT) is an approach to machine translation that uses neural networks to predict the likelihood of a sequence of words. The main functionality of NMT is based on the sequence to sequence (seq2seq) architecture, which is described in Section \ref{subsubsec:seq2seq}.

\subsubsection{Sequence to Sequence Model}
\label{subsubsec:seq2seq}
Sequence to Sequence learning is a concept in neural networks, that helps it to learn sequences. Essentially, it takes as input a sequence of tokens (words in our case)
\begin{equation*}
X=\{x\textsubscript{1}, x\textsubscript{2}, ..., x\textsubscript{n}\} 
\end{equation*}
and tries to generate the target sequence as output
\begin{equation*}
Y = \{y\textsubscript{1}, y\textsubscript{2}, ..., y\textsubscript{m}\}
\end{equation*}
where x\textsubscript{i} and y\textsubscript{i} are the input and target symbols respectively.

Sequence to Sequence architecture consists of two parts, an Encoder and a Decoder. 

The encoder takes a variable length sequence as input and encodes it into a fixed length vector, which is supposed to summarize its meaning and taking into account its context as well. A \textbf{L}ong \textbf{S}hort \textbf{T}erm \textbf{M}emory (LSTM) cell was used to achieve this. The uni-directional encoder reads the words of the Lithuanian texts, as a sequence from one end to the other (left to right in our case),
\begin{equation*}
\vec{h}\textsubscript{t} = \vec{f}\textsubscript{enc}(E\textsubscript{x}(x\textsubscript{t}),\vec{h}\textsubscript{t-1})
\end{equation*}
Here, E\textsubscript{x} is the input embedding lookup table (dictionary), $\vec{f}$\textsubscript{enc} is the transfer function for the LSTM recurrent unit. The cell state \textit{h} and context vector \textit{C} is constructed and is passed on to the decoder.

The decoder takes as input, the context vector \textit{C} and the cell state \textit{h} from the encoder, and computes the hidden state at time t as, 
\begin{equation*}
s\textsubscript{t} =  f\textsubscript{dec}(E\textsubscript{y}(y\textsubscript{t-1}), s\textsubscript{t-1}, c\textsubscript{t})
\end{equation*}

Subsequently, a parametric function out\textsubscript{k} returns the conditional probability using the next target symbol $k$. 
\begin{equation*}
(y\textsubscript{t}=k\mid y<{t}, X) = \frac{1}{Z}exp(out\textsubscript{k}(E\textsubscript{y}(y\textsubscript{t}-1), s\textsubscript{t}, c\textsubscript{t}))
\end{equation*}
$Z$ is  the normalizing constant, 
\begin{equation*}
\sum\textsubscript{j}exp(out\textsubscript{j}(E\textsubscript{y}(y\textsubscript{t}-1), s\textsubscript{t}, c\textsubscript{t}))
\end{equation*}
The entire model can be trained end-to-end by minimizing the log likelihood which is defined as
\begin{equation*}
L = -\frac{1}{N}\sum_{n=1}^{N}\sum_{t=1}^{T\textsubscript{y}\textsuperscript{n}}log p(y\textsubscript{t} = y\textsubscript{t}\textsuperscript{n}, y\textsubscript{<t}\textsuperscript{n}, X\textsuperscript{n})
\end{equation*}
where N is the number of sentence pairs, and X\textsuperscript{n} and y\textsubscript{t}\textsuperscript{n} are the input sentence and the t-th target symbol in the n-th pair respectively. 

The input to the decoder was one hot tensor (embeddings at word level) of 2,00,000 English sentences while the target data was identical, but with an offset of one time-step ahead.

\subsection{Architecture}
\begin{figure}[ht]
    \centering
    \includegraphics[scale=0.5]{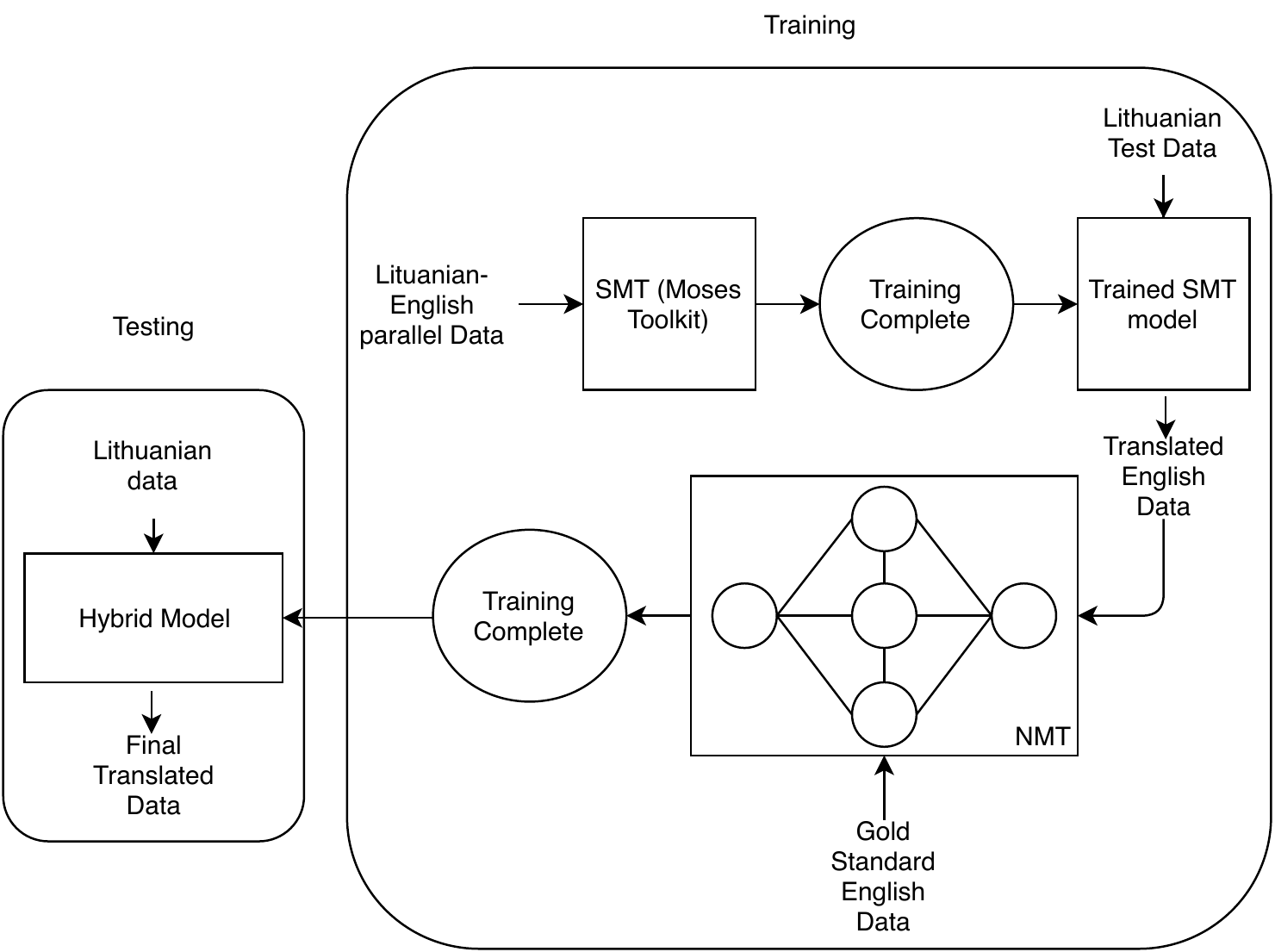}
    \caption{Architecture}
    \label{fig:my_label}
\end{figure}

\subsubsection{Training}
For the training purpose, 7,62,202 , preprocessed, Lituanian-English sentence pairs were fed to Moses Toolkit. This created a SMT translation model with Lithuanian as the source language and English as the target language. Thereafter, we had 2,00,000 Lithuanian-English sentence pairs, from which the Lithuanian sentences were given as input to the SMT model and it gave 2,00,000 translated English sentences as output. Now, this 2,00,000 translated English sentences and the respective gold standard 2,00,000 sentences, from the Lithuanian-English sentence pair, were given as input to a word embedding based NMT model. As a result, this constituted our Hybrid model. 

\subsubsection{Testing}
For the testing purpose, 10k Lithuanian Sentences were fed to the Hybrid model, and the output, when checked using BLEU \cite{papineni2002bleu}, resulted in an accuracy of 21.6. The training and testing architecture is shown in Figure \ref{fig:my_label}

\section{Results}
\label{sec:results}
WMT2019 provided us with a test set of Lithuanian sentences in .SGM
format. This file was parsed and fed to our hybrid system. The output file was again converted to .SGM format and submitted to the organizers. Our system garnered a BLEU Score of 17.6, when it was scored using automated accuracy metrics. Other accuracy scores are mentioned in Table \ref{table2}. \nocite{*}

\begin{table}[H]
\centering
\begin{tabular}{|c|c|}
\hline
\textbf{Metric} & \textbf{Score} \\ \hline
BLEU            & 17.6           \\ \hline
BLEU-cased      & 16.6           \\ \hline
TER             & 0.762          \\ \hline
BEER 2.0        & 0.497          \\ \hline
CharactTER      & 0.718          \\ \hline
\end{tabular}
\captionsetup{justification=centering}
\caption{Accuracy scores calculated using various autmoated evaluation metrics.}
\label{table2}
\end{table}

\section{Conclusion}
\label{sec:conclusion}
The paper presents the working of the translation system submitted to WMT 2019 News Translation shared task. We have used Word Embedding based NMT on top of SMT, for our proposed system. We have used a single LSTM layer as an encoder as well as a decoder. As a future prospect, we plan to use more LSTM layers in our model. We plan to create another model that incrementally trains both the SMT and NMT systems in a pipeline to improve the translation quality.

\section*{Acknowledgement}
The reported work is supported by Media Lab Asia, MeitY, Government of India, under the Visvesvaraya PhD Scheme for Electronics \& IT.

\bibliography{acl2019}
\bibliographystyle{acl_natbib}

\end{document}